%% file: arxiv_v1.tex
\documentclass{article}
\usepackage{arxiv}
\usepackage[round,authoryear]{natbib}

\usepackage{amsmath}
\usepackage{graphicx}
\usepackage{booktabs}
\usepackage[size=tiny]{todonotes}
\usepackage{makecell}
\usepackage{float}

\input{math_commands.tex}

\usepackage{hyperref}
\usepackage{url}

\title{Hypothesis-guided discovery of cognitive algorithms via program refinement}

\author{
Huiwen Alex Yang \\
Department of Psychology \\
University of California, Berkeley \\
\texttt{hw.yang@berkeley.edu}
\And
Mark K. Ho \\
Department of Psychology \\
New York University
\And
Bill D. Thompson \\
Department of Psychology \\
University of California, Berkeley
}

\begin{document}

\maketitle

\begin{abstract}
Developing cognitive models of algorithmic reasoning from behavioral data is a central problem in cognitive science that challenges current methods. Traditional approaches to cognitive modeling are interpretable and benefit from human expertise, but lack flexibility and scalability. Emerging techniques using large language models (LLMs) for de novo generation of cognitive models are scalable and flexible, but lack a role for human expertise and have mostly been applied to simpler tasks than algorithm recovery. We propose a hybrid system that treats discovery of cognitive algorithms as a program refinement problem. Human-created cognitive models are expressed as probabilistic programs and provided to a system of LLM agents with a mandate to: identify mismatches between model and behavior; propose code-level modifications within researcher-specified constraints; and verify structural fidelity. Revisions propagate to a probabilistic inference module that performs inference for latent variables and data likelihood computations. We evaluate the pipeline on human behavior in a problem-solving paradigm that exposes a variety of cognitive algorithms. Revised models consistently improve model fit relative to ancestral models and reveal a small set of recurring innovations that capture meaningful behavioral variability in this task.
\end{abstract}

\section{Introduction}
An overarching goal of modern cognitive science and artificial intelligence is to advance the science of cognitive modeling. Adaptive and scalable methods to induce interpretable cognitive models from behavioral data is a key frontier within this goal. In particular, advances that support induction of models for \emph{algorithmically-structured} behaviors would enable innovation across many sub-areas, and substantively expand the reach of cognitive science. Recent advances in LLM-supported modeling \citep[e.g.][]{rmus2025generating,curtis2025llm} establish new potential for scalable generation of cognitive models. However, current methods (1) generally underutilize human expertise; (2) often produce uninteresting models that reflect the LLM's prior; and (3) have mostly been applied at scale to simpler problems than algorithm recovery. We propose a solution and evaluate it in depth on a single well-suited problem-solving paradigm. Our approach views model discovery as \emph{program refinement}: a scientist's best hypothesized models are expressed as probabilistic programs and provided to a system of LLM agents with a mandate to identify targeted revisions that improve the fit between these initial proposals and human data; revised programs propagate to a probabilistic inference module that performs inference for latent variables and data likelihood computations. Our approach offers systematic integration of human expertise into LLM-supported discovery of algorithmic cognitive models.

\section{Background}
Standard approaches to cognitive modeling classically rely on model selection among a fixed set of models \citep{farrell2018computational}. This approach benefits from human expertise, strong interpretability, and normative standards for likelihood-based inference \citep{wilson2019ten}. It has been successful in domains such as multi-option choice or reward learning tasks, but involves fundamental limitations that become particularly apparent when applied to more complex settings; in particular, it is typically very challenging to capture the richness and variety of human thinking in \emph{algorithmically-structured} behaviors that unfold over time such as problem-solving \citep{allen2020rapid}, planning \citep{van2023expertise}, repeated games \citep{brockbank2024repeated}, or social interactions \citep{yu2024people,jha2025modeling}. For example, imagine observing the actions of a skilled (Rubik's) Cuber, and trying to reverse-engineer a cognitive model of their underlying knowledge: the model would need to specify how the person's actions depend not only on a complex set of cube features, but also potentially on a history of actions and their outcomes; in other words, the model would need to identify the structure of a latent \emph{cognitive algorithm}.       

At least two challenges make this a very difficult problem. One fundamental challenge is that the space of plausible cognitive algorithms is large and richly structured, making it difficult to anticipate relevant model variants. Another fundamental challenge is that even when a model is generally a good hypothesis, human behavior is often not well-captured by a single algorithm; even in domains where people learn canonical algorithms, such as mathematical reasoning \citep{siegler1984strategy}, people often exhibit systematic but idiosyncratic deviations from idealized procedures, including local execution errors, minor rule variations, and alternative tie-breaking conventions. Together, these challenges highlight a mismatch between classical approaches to model specification and the variability of real behavior. 

\section{Related Work}
In response to the limitations of traditional cognitive modeling, efforts have focused on model \emph{construction}. For example, an influential framework views model construction as \emph{program induction} \citep{ellis2023dreamcoder,jiang2024finding,piantadosi2016logical}. These methods typically specify a domain-specific language and perform statistical inference over expressible programs, extending a tradition of grammar-induction \citep{o2009fragment}, with successful application to, e.g., language \citep{yang2022one} and concept learning \citep{piantadosi2012bootstrapping,lake2015human,rule2024symbolic} but often require specification of complex languages and sampling algorithms. These difficulties motivated more general systems for program induction, driving development of probabilistic programming languages such as WebPPL \citep{dippl} that are natively enriched with probabilistic semantics and generic methods for statistical inference. Probabilistic programming frameworks resolve some difficulties of domain-specific methods, but remain primarily a tool for building hand-crafted models.       

In parallel, advances have been made in automated model discovery. For example, one approach generates models by training artificial neural networks on very large datasets \citep{agrawal2020scaling}. This approach has been examined in decision-based problems \citep{peterson2021using} as well as more algorithmic behaviors \citep{ji2025discovering, thompson2022complex}; it is scalable and can learn complex or counterintuitive models, but is not easily interpretable and is data-intensive. Prior theories can be integrated via initialization of network weights by pre-training on synthetic data generated from structured cognitive models \citep{bourgin2019cognitive}. More recently, emerging methods have leveraged the increasing capabilities of LLMs. One class of approaches simply fine-tunes LLMs on large volumes of human behavioral data \citep{binz2025foundation}; this is also data-intensive and the resulting models are difficult to interpret \citep{bowers2025centaur}. 

Another class of methods leverages the capabilities of LLMs for program synthesis \citep{ellisSynthesizingTheoriesHuman2022, pourcel2025self}. Under this approach, cognitive models can be expressed as programs in, e.g., Python. One stark proposal is to fully automate the larger enterprise of cognitive science \citep{jagadishCanWeAutomatize2026}. Recent approaches have used LLMs to directly propose cognitive models \citep{rmus2025generating}, integrating LLM proposals into model search or inference pipelines \citep{colas2025language}, using them to guide proposal generation \citep{curtis2025llm} or improve optimization over program spaces \citep{piriyakulkijDoingExperimentsRevising2024}, or leveraging natural language to bias synthesis toward more human-like solutions \citep{ellisHumanlikeFewShotLearning2023}. These approaches are most closely related to our framework, especially the approach of \cite{rmus2025generating}. 

However, these methods focus primarily on de novo generation of candidate programs -- a highly under-constrained problem that could benefit greatly from human expertise and the constraints of existing hypotheses or theories. In our experience (and commonly noted in the field), the programs proposed by automated discovery can often seem uninteresting or overly influenced by an LLM's prior over Python programs in general, being disconnected from the principles that researchers in the area would consider meaningful innovations, or e.g. simply enumerating contingencies. 

Another limitation of existing approaches is that they have primarily been applied to canonical problems that often lack rich algorithmic structure, such as multi-option decision-making \citep[e.g.][]{jagadish2026closing} or reward-learning tasks \citep{rmus2026generating} or judgment problems \citep{prystawski2026auto}. While there are many good reasons to begin with problems that have received much attention within cognitive science, this focus leaves open what we see as one of the most exciting opportunities presented by these methods: the opportunity to go \emph{beyond} the problems we have focused on historically in part as a reflection of limited methods. Recent advances in e.g. program synthesis \citep{curtis2025llm} and evolutionary coding \citep{tang2026alphaagentevo} suggest that LLM discovery of rich algorithmic structure can be very effective, and that this approach can be applied in behavioral contexts \citep{wang2026discovering}.

\section{Key Contribution}
We propose a system for \emph{hypothesis-guided} discovery of algorithmically structured cognitive models. Our approach views model discovery as a program-refinement problem. We express existing models as probabilistic programs, which are provided along with human data to a system of LLM agents that iteratively adapt the programs to better account for observed behavioral trajectories. LLM agents are instructed to identify mismatches between a program and behavior, and to propose targeted modifications as executable code. In particular, the agents are given semantically rich instructions in natural language, such that their revisions can be precisely scoped. Here we used this scoping to constrain the agents' modifications to preserve the basic structure of the original model. We represent this structure explicitly through structured annotated program components and revision specifications, which constrain how agents can introduce new decision rules, internal state, and other components. An additional benefit is that agents can explain their revisions in natural language. 

The pipeline consists of four modules that operate sequentially. Given a task specification, base model programs $\left(p_{t_1}, p_{t_2}, \ldots, p_{t_n}\right)$, and a behavioral trajectory $d$, the system produces and evaluates a revised program $\hat{p}_t$. Three modules, the Game, Coding, and Audit modules, are implemented using LLMs (in this paper: OpenAI's GPT-5.4). A Game module interprets the trajectory and generates a structured update signal along with a natural language rationale; a Coding module maps this signal to program modifications; an Audit module verifies correctness and fidelity. In contrast, the Inference module is not LLM-based; it instead performs probabilistic program evaluation using standard methods for probabilistic inference, computing the likelihood of the observed behavior under the revised program. We evaluate the system on human behavior in a novel problem-solving task that admits multiple strategies and induces diverse behavioral patterns across experimental conditions. The pipeline reliably produces revised programs that improve model fit while remaining consistent with the original strategy space.

\subsection{Cognitive models as probabilistic programs}
Following previous work in cognitive science~\citep{griffithsBayesianModelsCognition2024} and hierarchical reinforcement learning~\citep{andre2002state}, we use probabilistic programs as a representation for cognitive models that capture structured problem-solving strategies. Probabilistic programs combine the expressivity of structured programs with the flexibility of probabilistic models to provide a general (indeed, universal) representation for sequential decision-making~\citep{vandemeent2021introductionprobabilisticprogramming}. For example, a probabilistic program can represent a hierarchical policy in which a high-level controller probabilistically decides between executing different algorithms instantiated as low-level policies (e.g., turning on colors in a blue-green-red order versus green-blue-red order). The current work uses the FlipPy Python library~\citep{flippy-repo}, which takes a subset of Python as a ``host'' language and combines it with sampling and conditioning statements to express probabilistic programs. One of the benefits of using FlipPy, as opposed to other probabilistic programming languages such as Church~\citep{goodman2014churchlanguagegenerativemodels} (LISP-based) or WebPPL~\citep{dippl} (JavaScript-based), is that LLMs demonstrate strong biases to program in Python~\citep{twist2025llms}.

\subsection{Strategy Discovery and Evaluation System}
\begin{figure}[t!]
\begin{center}
\includegraphics[
    width=1.0\textwidth,
    trim={0 5 0 5},
    clip]{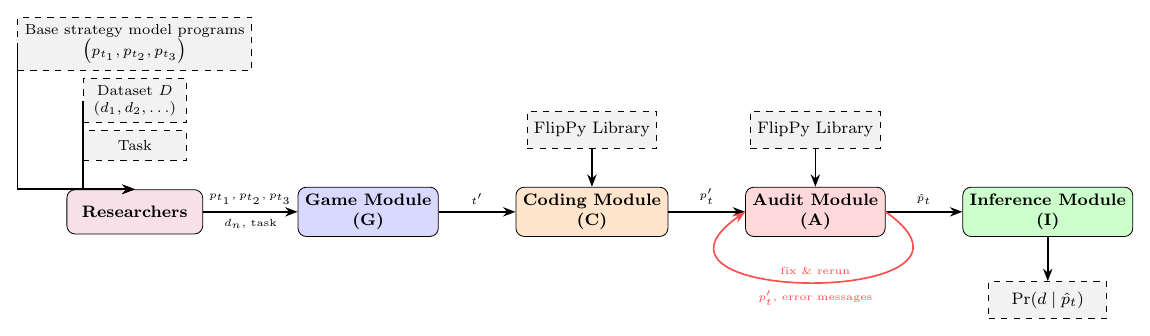}
\end{center}
\caption{Strategy discovery and evaluation system}
\label{figure-pipeline}
\end{figure}

Our system transforms base model programs into revised programs: given base model program $(p_{t})$, participant trajectory $d$, and a task specification, the system produces a revised program $\hat{p}_t$ and evaluates its likelihood $\Pr(d \mid \hat{p}_t)$. The system consists of four modules: the \textbf{Game Module} ($G$), \textbf{Coding Module} ($C$), \textbf{Audit Module} ($A$), and \textbf{Inference Module} ($I$). These modules operate in a structured sequence: each stage transforms the output of the previous one, jointly producing validated revised programs. The \textbf{Game Module} and \textbf{Coding Module} are separated to distinguish behavioral reasoning from code generation. The Game Module has access to the task specification and participant trajectories. The Coding Module specializes in translating these revisions into executable FlipPy code using the FlipPy library. This separation also reduces context size for each module, improving output reliability. Fig.~\ref{figure-pipeline} provides a high-level overview of this system.

\begin{enumerate}
\item \textbf{Researchers' input to the system.} The system is initialized with a set of human-written cognitive models as inputs: base strategy model programs $\left(p_{t_1}, p_{t_2}, p_{t_3}\right)$, participant data $d$ from dataset $D$, and the task specification. These inputs define the initial state of the system and are provided to the Game Module $G$.

\item \textbf{Game Module ($G$).} For every participant data $d$ and base strategy program $p_{t}$, the Game Module $G$ generates a strategy revision signal $t'$, representing a structured transformation of the base strategy program class induced by the observed behavior from this participant. The revision signal specifies targeted modifications to the decision policy, including rule-level changes, added latent variables and parameters, and operational constraints that preserve the original program structure while introducing behaviorally motivated revisions.
    
\item \textbf{Coding Module ($C$).} Conditioned on $t'$ and the FlipPy library, $C$ implements a program synthesis step that maps the updated strategy specification to an executable program $p'_t$.
    
\item \textbf{Audit Module ($A$).} The audit module $A$ performs verification over $p'_t$ under the FlipPy runtime environment. It checks for syntactic and runtime correctness. The output is a validated program $\hat{p}_t$. In case of failure, $A$ applies a bounded repair procedure using execution error signals and re-verification.
    
\item \textbf{Inference Module ($I$).} The inference module evaluates $\hat{p}_t$ on participant data $d$ and computes the likelihood $\Pr(d \mid \hat{p}_t)$ under the revised strategy model.
\end{enumerate}

\subsection{Generation, selection, and validation of revised strategies}
We ran the full system twice per participant per base strategy model, producing two independently generated revised programs for each participant-base strategy pair. Programs produced by the Coding Module ($C$) were passed through the Audit Module ($A$), which allowed up to five iterative repair attempts to resolve syntactic or runtime errors. Programs that remained non-executable were excluded from subsequent analyses. The resulting audited programs constitute the candidate revised-strategy set used in subsequent evaluation. Because different runs can produce different revisions of the same base strategy, we subsequently grouped candidate programs into broader classes of qualitatively similar revisions. This clustering and selection procedure was applied separately within each base-strategy family to obtain a compact set of representative revised strategy models for evaluation. After generating and selecting the representative revised strategy models, we evaluated whether they provided improved fit to participants' behavioral data. We first evaluated the representative programs on the original dataset and compared their fit with that of the corresponding base strategy models. We then conducted the same evaluation on a held-out set of trials that was not used during program generation or selection, allowing us to assess whether improvements in fit generalized beyond the trajectories used to develop the revised strategies.

\subsection{Human problem-solving algorithms}
We applied our pipeline to human behavior in a problem-solving paradigm that examines how people develop algorithmic strategies to solve structured tasks \citep{yang2026strategy}. The dataset contains trial-level data from 150 participants across multiple task environments, providing an ideal testbed because participants developed a variety of strategies, producing heterogeneous behavior. Some participants' behavior is well captured by existing cognitive models of varying structural complexity, whereas other participants exhibit behavior that is not adequately captured by these models, providing an opportunity for our pipeline to discover more accurate algorithms.

\subsubsection{Task Structure}
\begin{figure}[]
\begin{center}
\includegraphics[width=1.0\textwidth]{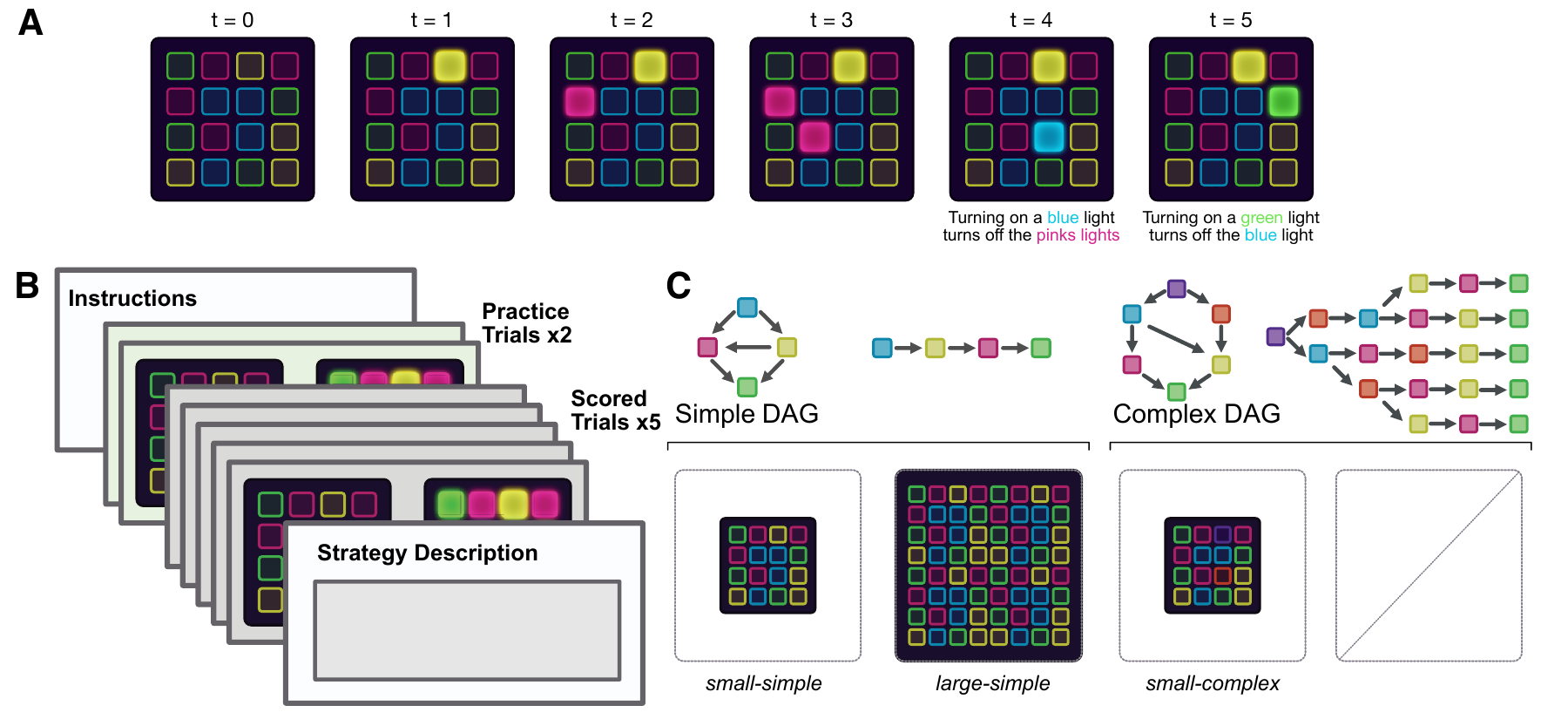}
\end{center}
\caption{Behavioral experiment. (\textbf{A}) Task Structure: first five actions in an example trial. (\textbf{B}) Procedure: instructions, two practice trials, and five scored trials, followed by a strategy description. (\textbf{C}) Experimental conditions defined by board size (small vs. large) and task structure (simple vs. complex DAG), yielding three conditions. Representative grids and DAGs for each condition, along with a set of working orderings.}
\label{figure-design}
\end{figure}

Participants completed a problem-solving task involving a grid of colored lights (Fig.~\ref{figure-design}A). The goal is simple: turn on all the lights. Each light can be toggled by clicking. Clicking a light could, however, turn off other lights that were already on. The task was defined by a set of colors and their dependency relations, where a dependency from color \(i\) to color \(j\) meant that clicking any light of color \(i\) turned off all currently active lights of color \(j\). These dependencies formed a directed acyclic graph (DAG) -- e.g. blues turn off pinks, greens turn off blues. Grid size, number of colors, and the DAG structure varied across conditions. Participants were not told the underlying dependencies and had to discover an effective way to solve the task. See \citep{yang2026strategy} for further detail.

Under these dynamics, the task can be solved by activating colors in an order consistent with the DAG. Specifically, any \emph{topological sort} of the DAG, a permutation of its colors in which every source color precedes the colors it turns off, defines a valid color ordering. For example, if blue and yellow both turn off green, either blue--yellow--green or yellow--blue--green is a valid ordering. Participants therefore did not necessarily need to recover the full DAG; identifying a valid ordering was sufficient to solve the task efficiently. In the studies reported here, participants are not informed of the DAG's structure: they must find a way to solve the problem (turn on all the lights) efficiently, which could involve trying to establish the structure and using that knowledge to complete the task, or simply turning on lights at random, or any one of innumerable algorithms between these extremes; our goal as modelers is to characterize the structure of the algorithms developed by participants.

\subsection{Base Strategy Models}
\label{section-base_strategy_models}
We begin with the 3 strategy models developed in the original study to characterize participants' behavior in this task \citep{yang2026strategy}. These models were specified \emph{a priori} to capture qualitatively distinct ways of approaching the task, ranging from surface-level spatial heuristics to strategies that infer and exploit the task's latent structure. In the present work, these models serve as the starting points for program refinement. Each model has a fixed program structure with latent parameters inferred during model fitting, and we implemented them using the probabilistic programming library FlipPy \citep{flippy-repo}.

\paragraph{Spatial (SP)} Lights are turned on using simple spatial patterns: e.g. row-by-row, column-by-column, or back-and-forth zigzag sequences. Pattern and lapse rate are latent parameters inferred during model fitting. Does not consider color or turn-off relations. This strategy reflects a failure to recognize or utilize the relevance of color-based structure altogether, instead relying on generic spatial heuristics that would be effective in tasks without latent dependencies.

\paragraph{Color Blocking (CB)} Systematically turns on all lights of one color before moving to the next, without considering dependencies between colors. The color order and a probabilistic lapse rate are treated as latent parameters inferred during model fitting. This strategy reflects a partial discovery of task structure: participants recognize that color is a relevant organizing feature and exploit this regularity, but do not infer or utilize the underlying dependency relationships between colors.

\paragraph{Hierarchical Two-Phase Filling (TP)}
This model represents a hierarchical strategy in which a valid topological ordering is first inferred and then used to complete the board. In phase one, the model explores lights across colors, only testing lights not yet activated, thereby accumulating constraints on the set of possible orderings. In phase two, it commits to a single ordering drawn from the set of topological orders still consistent with the observed turn-offs. For example, if a yellow light is observed to turn off a green light, then any ordering placing yellow before green is eliminated from the feasible set. As more such constraints are observed in phase one, the set of admissible orderings becomes increasingly restricted, making correct orderings more likely. This strategy captures participants who infer color dependencies and attempt to exploit them in a heuristic, potentially imperfect manner. Two latent parameters are inferred: the transition point between phases and the phase-two ordering used to complete the board.

\section{Results}
\begin{figure}[t]
\begin{center}
\includegraphics[width=1.0\textwidth]{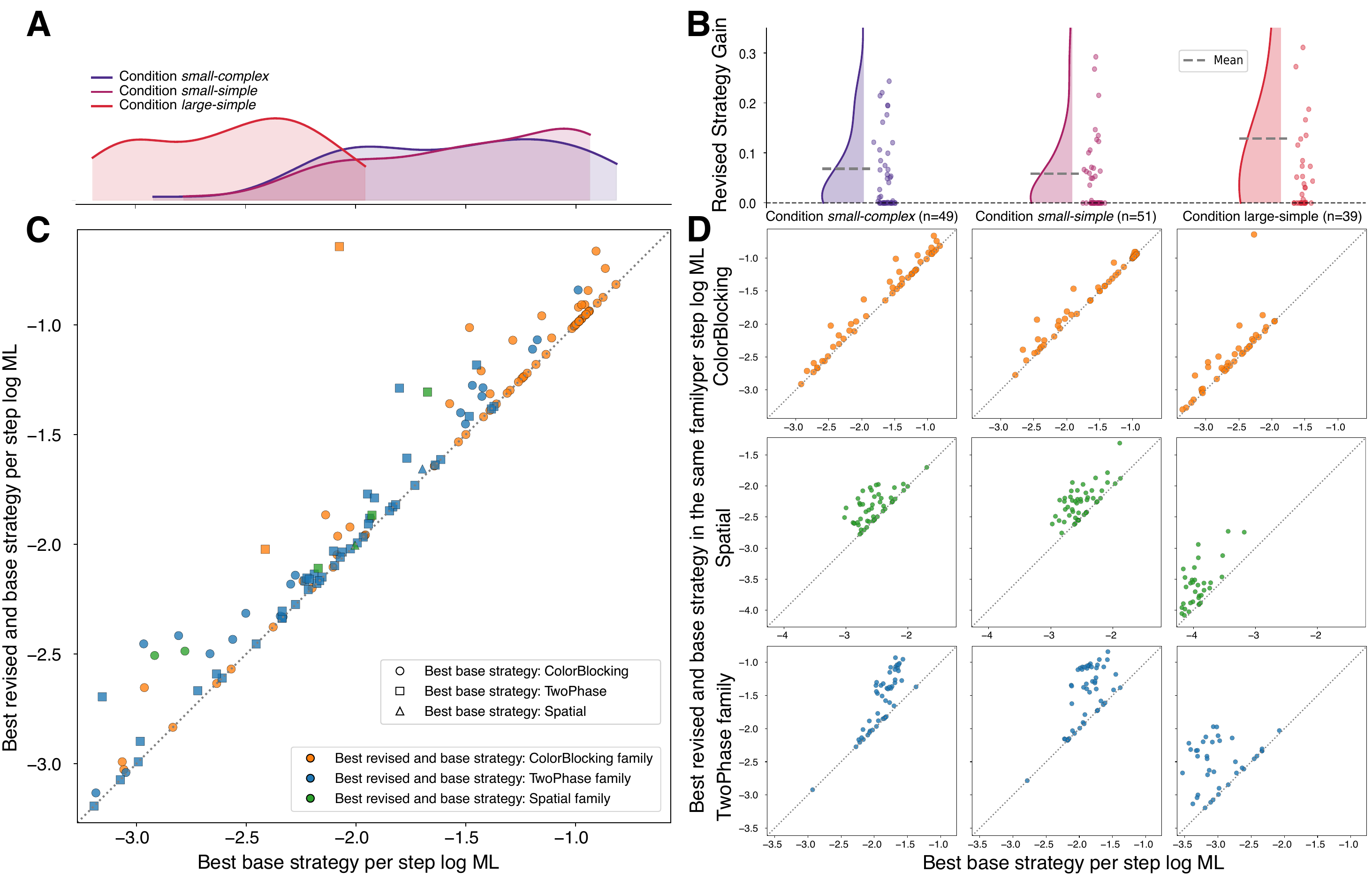}
\end{center}
\caption{Base strategies provide adequate baseline fit, and revised strategies yield improved model fit for a substantial subset of participants. (\textbf{A}) Kernel density estimates of per-step log marginal likelihood (log ML) for the best-fitting base strategy, shown by condition (x-axis shared with panel C). (\textbf{B}) Improvement from revised strategies by condition, shown as per-step log ML gain (gain = revised - base). Half-violins show the distribution, points show individual participants, and dashed lines indicate condition means. The y-axis is truncated to [0, 0.35] for clarity (9/139 participants have higher gain and exceed this range). (\textbf{C}) Participant-level comparison of best base vs. best overall (base or revised) fit; points above the diagonal indicate improvement. Color denotes the family of the best revised strategy; shape denotes the best base strategy. (\textbf{D}) Same comparison as (C), broken down by strategy family and condition, with the y-axis showing the best revised variant within each family.}
\label{figure-improvement}
\end{figure}

\subsection{Baseline Model Fit and Benchmark Quality}
To establish a benchmark for evaluating refined programs, we first assessed fit for three base models. We computed data likelihood for each model on the final correctly completed trial for each participant. Participants who did not complete any trial correctly were excluded, leaving 49 trials in \emph{small-complex}, 51 trials in \emph{small-simple}, and 41 trials in \emph{large-simple}. Tab.~\ref{table-base_fit_summary} (Appendix~\ref{appendix-base}) summarizes trial-level model fit using per-step log marginal likelihood. The baseline models produced numerically stable and broadly computable fits (finite-fit coverage: 95–100\% across conditions), with relatively tight bootstrap confidence intervals around condition-level means and medians. Fig.~\ref{figure-improvement}A shows the distribution of best-fitting base-strategy log ML across conditions. The results reveal meaningful variation in fit across participants: some are well explained by the base strategies, while others are less well captured, leaving room for improvement through model refinement.

\subsection{Discovery of Revised Strategies}
We first generated candidate revisions by running the pipeline twice for each participant and each of the 3 base strategy models. This produced 846 candidate revised programs in total. The Audit Module validated 733 programs (86.6\%); we excluded the rest because they remained non-executable after the allowed repair attempts. Within each base-strategy family, we clustered audited revisions according to the semantic similarity of their natural-language revision proposals. The largest clusters captured recurring types of modifications to the corresponding base strategies (Table~\ref{table-cluster}).

Across the 3 base strategy families, we identified 8 recurring revision clusters: 3 within Color Blocking, 2 within Spatial, and 3 within Two-Phase. Cluster size varied substantially, with some revisions recurring across many participants and others appearing more rarely. We selected 3 representative audited programs closest to the centroid of each cluster, yielding 24 prototypical revised models for the subsequent analyses.

\subsection{Improvement in Model Fit from Revised Models}
Fig.~\ref{figure-improvement}~B reports the improvement in per-step log marginal likelihood (log ML) for each participant. The improvement is computed as the difference between the per-step log ML of the best-fitting strategy (selected over both base and revised strategies) and the per-step log ML of the best-fitting base strategy alone. For a substantial subset of participants 53.2\% (74/139), the inclusion of revised strategies yields meaningful improvements in fit. This indicates that, for these participants, the best-explaining model is not contained within the original base strategy set, but instead lies within the space of revised programs.

Fig.~\ref{figure-improvement}C,D further characterize these gains. Panel C shows that for 53.2\% (74/139) of participants, a revised strategy provides the overall best fit. Panel D complements this analysis by evaluating improvements within each strategy family, rather than only the best-fitting strategy. As a result, improvements are observed across all families (ColorBlocking: 54\%; Spatial: 76.9\%; TwoPhase: 80.4\%), including cases where a revised variant improves the fit of a non-winning base model without changing the participant’s overall best-fitting model. Together, these results demonstrate that the program refinement system reliably generates structured variants of base models that better account for observed behavior across participants.

\begin{figure}[]
\begin{center}
\includegraphics[
    width=1.0\textwidth,
    trim={0 20 0 20},
    clip]{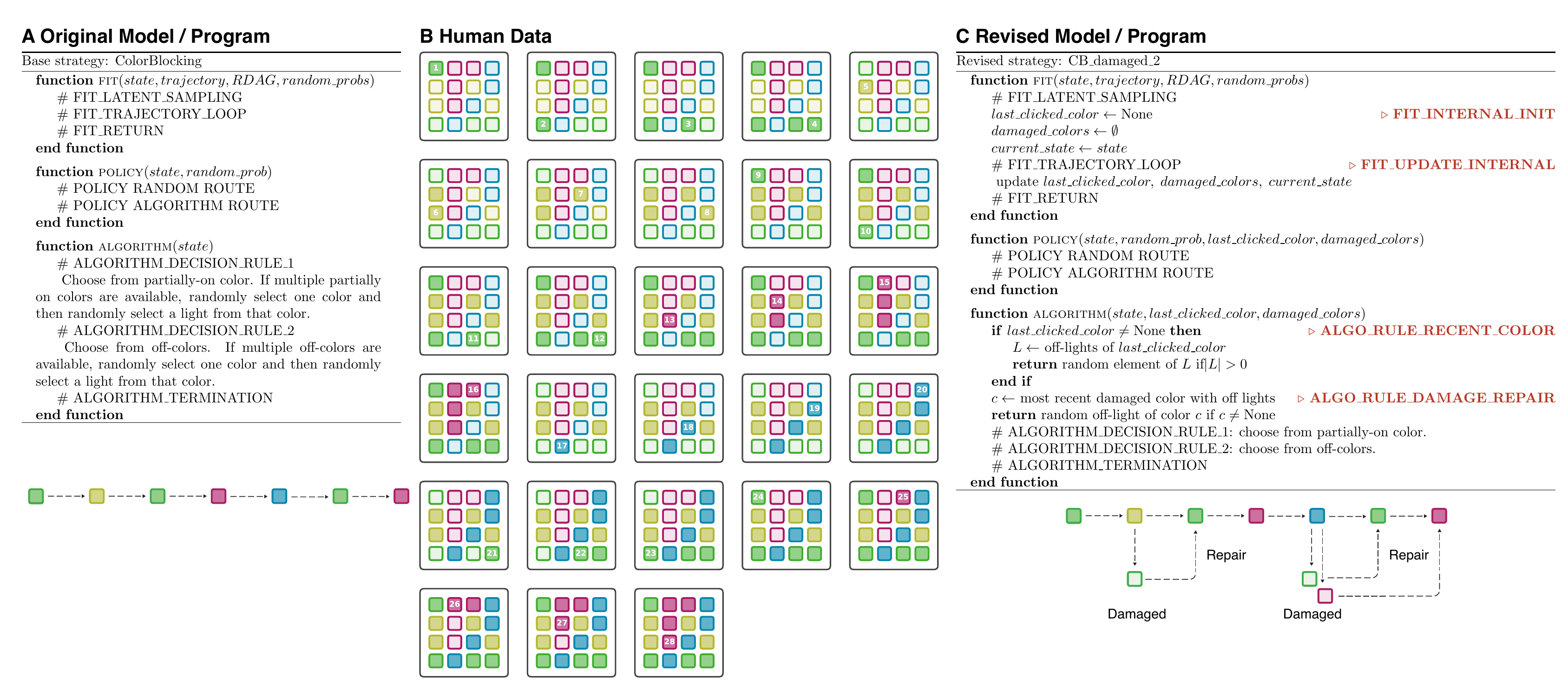}
\end{center}
\caption{A representative case of a participant who is reasonably well fit by the base ColorBlocking strategy but exhibits improved fit under a revised variant. (\textbf{A}) Core logic of the base CB strategy, which predicts the behavior to complete one color at a time. (\textbf{B}) Participant behavior data. (\textbf{C}) Core logic of the revised strategy variant \texttt{CB\_damaged\_2}. Annotations highlight key modifications to the base strategy. Modifications capture the participant’s behavior to repair previously damaged colors.}
\label{figure-example}
\end{figure}

Fig.~\ref{figure-example} illustrates a representative case. This participant is reasonably well fit by the base ColorBlocking (CB) algorithm, but exhibits improved fit under a revised variant, \texttt{CB\_damaged\_2}. While the base CB algorithm captures the participant’s tendency to complete one color at a time, it fails to account for behavior following a \emph{damage event} (i.e. a turn-off event, where previously-on lights are turned off by an action that violates the DAG). Specifically, after observing that activating a yellow light turns off the greens, the participant completes the yellow lights and then immediately returns to repair the previously `damaged' green lights. The revised strategy \texttt{CB\_damaged\_2} captures this damage-aware repair behavior, resulting in improved fit for this participant by proposing a meaningful qualitative innovation to the base algorithm. 

\subsection{Generalization and synthetic recovery}
To assess whether these improvements generalize beyond the data used for model refinement, we additionally evaluated the 24 prototypical revised strategy programs on the held-out dataset. In the held-out data, 52.6\% (72/137) of trials showed improved fit under a revised program. Thus, the improvement observed during refinement was not restricted to the data used to generate and select revisions. The persistence of these improvements on held-out data suggests that the revised programs capture behavioral regularities that extend beyond the trajectories on which they were developed. Details on held-out analyses are provided in Appendix~\ref{appendix-heldout}.

As a complementary synthetic recovery test, we generated synthetic behavioral trajectories from the \texttt{CB\_damaged\_2} strategy and applied our pipeline to individual trajectories. Of the 215 valid revised programs generated by our pipeline, 111 (51.6\%) achieved a higher mean held-out log likelihood across 40 held-out trials than the base Color Blocking program, while 18 (8.4\%) recovered at least 75\% of the predictive gap between the two models. 104 (48.4\%) did not achieve an improvement in mean held-out log likelihood over the base model over held-out trials. The variability in recovery may in part reflect that the pipeline sometimes produces revised programs that overfit to idiosyncratic features of the single behavioral trajectory used to generate each revision, rather than recovering the underlying strategy. Overall, these results provide evidence that the procedure can recover predictive consequences of a richer strategy, while also illustrating the limitations of trajectory-level strategy modeling. Further synthetic recovery analyses are provided in Appendix~\ref{appendix-synthetic}.

\subsection{Revised models reveal a small set of recurring algorithmic innovations}
To better understand the structure of the revisions produced by the system, we analyze their organization in semantic space. Clustering identifies recurring themes and yields a compact set of representative variants; Tab.~\ref{table-cluster} reveals two robust regularities in the structure of revisions. First, revisions are highly clustered in semantic space: across 462 total revision proposals, behavior concentrates into a small number of dominant groups (6 clusters for ColorBlocking, 3 for Spatial, and 6 for TwoPhase), with a small subset of clusters accounting for the majority of revisions within each strategy family. Second, the dominant clusters are strikingly consistent across strategy families and reflect an unanticipated feature: \emph{locally grounded behavioral adjustments}. Themes related to damage repair, local corrective actions, and deterministic tie-breaking appear as the largest cluster(s) in all three strategy families. These account for approximately 73\% of ColorBlocking revisions (161/222), 67\% of Spatial revisions (8/12), and 79\% of TwoPhase revisions (180/228).

\begin{table}[t]
\caption{Top three revision clusters per strategy family, ranked by size (\# revisions within the cluster). Cluster descriptions are LLM-generated for interpretation only and are not used in modeling or analysis. Full results are in Appendix~\ref{appendix-cluster}.}
\centering
\begin{tabular}{lcl}
\toprule
\textbf{Strategy Family} & \textbf{Size} & \textbf{LLM-generated cluster description} \\
\midrule

ColorBlocking & 161 & Prioritize recent-damage repair before resuming default strategy \\
ColorBlocking & 44  & Damage-aware color-blocking with flexible partial completion \\
ColorBlocking & 8   & Within-color tie-breaking with consistent index order \\

\midrule

Spatial & 8  & Spatial scanning with temporary damage-repair mode \\
Spatial & 3  & Probabilistic local repair and repeat after events \\

\midrule

TwoPhase & 106 & Two-phase damage-aware repair and priority policy \\
TwoPhase & 74  & Short-term repair prioritization after observed damage \\
TwoPhase & 29  & Stateful color-focused exploration with immediate repair \\

\bottomrule
\end{tabular}

\label{table-cluster}
\end{table}

\section{Limitations and future work}
Our analysis focused on a single problem-solving paradigm, using just a single iteration of the propose-revise loop and a single underlying LLM. This \emph{narrow and deep} approach revealed important new findings relative to \emph{broad and shallow} approaches, but it is clearly important to explore whether other algorithmically-structured problem settings reveal further insight and to explore potential gains from alternative models. In future work, we envisage extending our approach to a larger multi-agent process or other canonical sampling algorithms such as MCMC or SMC procedures, and further exploring aspects of the harness that could support improved exploration in program space. Perhaps the most important issue not addressed here or in other work concerns the potential for program revisions to over-fit individual participants, trials, or tasks: methods for effective partial pooling and program compression across contexts are a clear priority for future work. Similarly, no current approach (including ours) offers a systematic way to assess how discovered programs relate to broader theoretical frameworks beyond lower-level procedural constraints; improved data-likelihood alone is an insufficient indicator of meaningful meaningful innovation, which here we assessed only heuristically--this broader question will be central as the area advances.

\section{Conclusion}
We proposed a program refinement framework for cognitive modeling that combines structured probabilistic programs with an LLM-based program revision system. We applied this system to the more challenging domain of algorithmically-structured sequential decision-making models. By enabling targeted revisions while preserving hypothesized structure and interpretability, our framework offers one principled way to integrate the increasing capabilities of frontier LLMs with (1) human expertise and (2) probabilistic statistical inference, and shows that these methods successfully extend to recovery of algorithmic models, a key frontier in cognitive science.

\newpage

\section*{Ethics statement}
The experimental protocol was reviewed and approved by the institutional review board (IRB) at our institution, and all participants provided informed consent prior to participation. Participants received a base payment of \$3 and a bonus of \$0.4 per correctly completed trial.

\section*{Reproducibility statement}
All code, data, and experimental materials associated with this work are available in the supplementary materials and in the accompanying anonymous OSF repository. The main paper and appendix provide detailed descriptions of the model refinement pipeline, base and revised strategy programs, model-fitting and inference procedures, and held-out evaluation. The appendix further provides complete descriptions of the revision clusters and additional analyses. The OSF repository contains the source code, experimental data, model programs, and materials necessary to reproduce the analyses and results reported in this work.

Code, data, and experimental materials associated with this work are not included with this arXiv version.

\newpage
\bibliographystyle{plainnat}
\bibliography{references,additional_references}

\newpage
\appendix

\section{Participant Instructions}

All participants received the same instructions prior to beginning the task, reproduced below:

\begin{quote}
On each trial, you will play a game. You will see a grid of lights like the one below.

Your goal is simple: turn on all the lights!

Every square on the grid is a light. You can turn a light on or off by clicking on it.

You will complete 2 practice trial(s). In these trials, every trial is the same game.

After you complete the practice trials, you will complete 5 scored trial(s). In these trials, every trial will be a different game.

You will get a bonus payment for every trial you complete (\$bonus per trial).

If you really cannot complete a trial, you can skip it (but you will not receive the bonus for that trial).

Good luck!
\end{quote}

\section{Base strategy model fit summary}
\label{appendix-base}
Table~\ref{table-base_fit_summary} summarizes the performance of the best-fitting base strategy model across conditions using per-step log-likelihoods. Overall, the base models produce stable and well-defined fits, with high finite-fit coverage (95–100\%) across all conditions, indicating that inference is numerically reliable for the vast majority of trials. The relatively tight bootstrap confidence intervals further suggest that condition-level estimates are well-constrained. There are also systematic differences across conditions: Conditions \emph{small-complex} and \emph{small-simple} exhibit similar fit quality, whereas Condition \emph{large-simple} shows substantially more negative per-step log-likelihood. This pattern is consistent with the increased trajectory length in Condition \emph{large-simple}, reflecting the increased task scale and associated demands. Taken together, these results indicate that the base models provide a stable but imperfect account of behavior, establishing a meaningful baseline against which improvements from revised strategies can be evaluated.
\begin{table}[H]
\centering
\caption{Descriptive summary of best-fitting base model performance by condition. Values reported here are per-step log-likelihoods, computed as the log marginal likelihoods of the best-fitting base strategy model normalized by trajectory length. 95\% confidence intervals are bootstrap percentile intervals ($B = 2000$).}
\label{table-base_fit_summary}
\begin{tabular}{lrrrrrrr}
\toprule
Condition & $N$ & Finite fits (\%) & Mean [CI$_{95}$] & SD & Avg.\ traj.\ length \\
\midrule
\emph{small-complex} & 49 & 100.0 & $-1.544$ [$-1.680$, $-1.414$] & $0.489$ & $39.7$ \\
\emph{small-simple} & 51 & 100.0 & $-1.464$ [$-1.598$, $-1.335$] & $0.479$ & $38.4$ \\
\emph{large-simple} & 41 & 95.1  & $-2.577$ [$-2.693$, $-2.465$] & $0.373$ & $125.6$ \\
\bottomrule
\end{tabular}
\end{table}

\section{Clusters of revisions}
\label{appendix-cluster}
Table~\ref{table-cluster_all} provides a comprehensive summary of revision clusters across all strategy families. Several regularities emerge. First, revisions are highly concentrated: within each strategy family, a small number of dominant clusters account for the majority of proposals. For example, in the ColorBlocking family, the largest cluster (size 161) substantially outweighs all others, and a similar skew is observed in the TwoPhase family, where the top two clusters (sizes 106 and 74) dominate. In contrast, the Spatial family contains fewer and smaller clusters overall, reflecting both lower revision volume and a more compact set of variation patterns.

Second, the dominant clusters across families reflect closely related functional themes. In particular, revisions centered on damage detection and repair, as well as short-term prioritization of recently affected lights, appear as the largest clusters in all three families. Deterministic tie-breaking and ordering strategies also recur, though typically at smaller scales. This suggests that, despite differences in the underlying base strategies, revisions are organized around a shared set of locally grounded behavioral adjustments.
\begin{table}[H]
\centering
\caption{Summary of revision clusters across strategy families. For each family, all clusters, ranked by size (number of revisions in the cluster), are shown. Descriptions are LLM-generated for interpretive purposes and are not used in the system or analysis.}
\label{table-cluster_all}

\begin{tabular}{c p{4.2cm} p{8.5cm}}
\toprule
\textbf{Size} & \textbf{Strategy Family} 
              & \textnormal{LLM-generated cluster description} \\
\midrule

\multicolumn{3}{c}{\textbf{ColorBlocking Strategy Family}} \\
\midrule

161 & Prioritize recent-damage repair before resuming default strategy &
After a turn-off/damage event, temporarily focus on repairing or avoiding the most recently damaged color(s) using a stable within-color click order, then resume the default color-blocking policy. \\
\midrule
44 & Damage-aware color-blocking with flexible partial completion &
A damage-aware color-blocking strategy that maintains memory of recently damaged colors and flexibly switches between continuing a color block and repairing or reordering priorities. \\
\midrule
8 & Within-color tie-breaking with consistent index order &
When selecting within a color, ties among same-color OFF lights are resolved using a deterministic spatial/index sweep instead of random selection. \\
\midrule
4 & Occasional post-completion or late-stage exploratory actions &
Rare exploratory deviations from strict termination, such as additional clicks after completion or near-completion, to capture behavioral variability. \\
\midrule
3 & Selection stickiness --- persist with chosen color until ON\_COLOR &
A persistence bias that keeps selecting the same color until it is fully ON, only switching when external state changes require it. \\
\midrule
2 & Optional exploratory lapse augmenting ColorBlocking policy &
A low-probability lapse mechanism that occasionally overrides the main policy, producing rare exploratory or erroneous clicks. \\
\midrule\midrule

\multicolumn{3}{c}{\textbf{Spatial Strategy Family}} \\
\midrule

8 & Spatial scanning with temporary damage-repair mode &
Spatial scanning with a conditional repair mode that activates after damage events, prioritizing restoration before resuming scanning. \\
\midrule
3 & Probabilistic local repair and repeat after events &
A stochastic mechanism that occasionally repeats or locally repairs recent actions after specific outcome events. \\
\midrule
1 & Structured-prior bias toward deterministic ordering &
A strong bias toward deterministic spatial ordering and reduced randomness in action selection. \\
\midrule\midrule

\multicolumn{3}{c}{\textbf{TwoPhase Strategy Family}} \\
\midrule

106 & Two-phase damage-aware repair and priority policy &
A two-phase strategy that prioritizes repairing damaged colors before proceeding with the exploit phase. \\
\midrule
74 & Short-term repair prioritization after observed damage &
Temporary prioritization of recently damaged colors before returning to the main policy. \\
\midrule
29 & Stateful color-focused exploration with immediate repair &
Stateful exploration that maintains color focus while inserting immediate repair steps when damage is observed. \\
\midrule
15 & Deterministic spatial tie-breaking within color selection &
Deterministic spatial ordering replaces random selection within color choices during execution. \\
\midrule
2 & Turn-off-triggered repair prioritization in exploitation &
Exploitation phase is temporarily interrupted after damage events to prioritize repair. \\
\midrule
2 & Deterministic within-color spatial ordering strategy &
A deterministic within-color ordering rule applied during the exploit phase. \\

\bottomrule
\end{tabular}
\end{table}

\section{Heterogeneity of best-fitting revised strategy implementations}
\label{appendix-revised}
Table~\ref{table-best_revised_stats} shows how best-fitting revised strategies are distributed across strategy families, clusters, and specific variants. Several patterns emerge. First, even within a given cluster, multiple variants are selected across participants, indicating that no single implementation uniformly dominates. For example, within the TwoPhase “damaged” cluster, all 3 variants are preferred by some participants, and similar splits appear in ColorBlocking clusters. This provides direct evidence of within-cluster heterogeneity in how shared high-level revision ideas are instantiated. Second, while certain variants account for a larger fraction of participants (e.g., TP\_repair\_1 with 22/75), the overall distribution remains diffuse, with a substantial portion of participants assigned to lower-frequency variants. This long-tailed pattern indicates that model fit is distributed across many implementations rather than concentrated on a single dominant strategy. Finally, these patterns are consistent across strategy families. In all three families, clusters such as repair and damage handling contain multiple viable variants, reinforcing the idea that common functional theoretical revisions admit multiple realizations on a program level. Together, these results support the conclusion that, although revisions are organized around shared cluster-level principles, the specific implementations that best explain behavior vary systematically across participants.
\begin{table}[H]
\caption{Number of participants best-fitted by each revised strategy, grouped by base strategy family, cluster type, and specific variant (N = 75).}
\label{table-best_revised_stats}
\centering
\begin{tabular}{cllcc}
\toprule
\textbf{Strategy Family} & \textbf{Cluster} & \textbf{Variant} & \textbf{Count} & \textbf{\%} \\
\midrule

\multicolumn{5}{c}{\textbf{TwoPhase (TP)}} \\
\midrule
TP & repair  & TP\_repair\_1  & 22 & 29.3\% \\
   &         & TP\_repair\_3  & 2  & 2.7\% \\
\midrule
TP & damaged & TP\_damaged\_2 & 9  & 12.0\% \\
   &         & TP\_damaged\_3 & 3  & 4.0\% \\
   &         & TP\_damaged\_1 & 3  & 4.0\% \\
\midrule
TP & local   & TP\_local\_1   & 7  & 9.3\% \\
\midrule

\multicolumn{5}{c}{\textbf{ColorBlocking (CB)}} \\
\midrule
CB & damaged & CB\_damaged\_2 & 11 & 14.7\% \\
   &         & CB\_damaged\_1 & 3  & 4.0\% \\
\midrule
CB & repair  & CB\_repair\_3  & 2  & 2.7\% \\
   &         & CB\_repair\_1  & 2  & 2.7\% \\
\midrule
CB & index   & CB\_index\_3   & 5  & 6.7\% \\
   &         & CB\_index\_1   & 1  & 1.3\% \\
\midrule

\multicolumn{5}{c}{\textbf{Spatial (SP)}} \\
\midrule
SP & repair  & SP\_repair\_3  & 2  & 2.7\% \\
   &         & SP\_repair\_1  & 2  & 2.7\% \\
\midrule
SP & damaged & SP\_damaged\_3 & 1  & 1.3\% \\
\midrule

\bottomrule
\end{tabular}

\end{table}

\section{Heldout Analyses}
\label{appendix-heldout}
\subsection{Heldout dataset}
Each participant completed five scored trials in the experiment. In the model refinement process and analyses reported in the main text, we used only the final correctly completed trial from each participant. The held-out dataset consists of the second-to-last correctly completed trial for each participant. Participants who did not complete any trial correctly, or who had only one correctly completed trial, were therefore excluded from the held-out analysis. This resulted in 49 trials in the \emph{small-complex} condition, 50 trials in the \emph{small-simple} condition, and 39 trials in the \emph{large-simple} condition.

\subsection{Held-out Evaluation}
We evaluated the held-out trials using the same model-fitting procedure as in the main analysis. First, we fit each of the 138 held-out trials to the 3 base strategies and selected the base strategy with the highest per-step log marginal likelihood for each trial. We then fit the same trials to the 24 prototypical revised strategies identified from the 3 strategy families and selected the strategy with the highest per-step log marginal likelihood among all base and revised strategies.

One held-out trial could not be evaluated because all 27 strategies produced a log marginal likelihood of $-\infty$. This trial contained 270 behavioral steps, but we could not determine whether the non-finite likelihoods were attributable to sequence length or to another numerical or model-specific issue. We therefore excluded this trial from the comparison, leaving 137 evaluable held-out trials.

For each held-out trial, we quantified improvement as the difference between the per-step log marginal likelihood of the best-fitting strategy among the base and revised strategies and that of the best-fitting base strategy alone. A positive value therefore indicates that at least one revised strategy provides a better fit to the held-out trial than any of the three base strategies. Across the 137 held-out trials, 72 (52.6\%) showed positive improvement under a revised strategy.

Similar to the main analysis, held-out improvements are observed across all three strategy families  (ColorBlocking: 52.6\%; Spatial: 86.7\%; TwoPhase: 83.2\%), including cases in which a revised variant improves fit relative to the corresponding base strategy without becoming the best-fitting strategy overall. 

\section{Synthetic Analyses}
\label{appendix-synthetic}

\subsection{Synthetic Dataset}
To evaluate recovery in a setting where the underlying behavioral strategy is known, we generated a synthetic dataset from the \texttt{CB\_damaged\_2} strategy. This strategy variant was selected because the original behavioral study identified damage-repair behavior as a common extension of the ColorBlocking strategy, with participants often returning to repair colors that had been damaged by subsequent actions \citep{yang2026strategy}. Thus, \texttt{CB\_damaged\_2} provides a concrete example of a behavioral regularity observed in participants and allows us to ask whether the discovery pipeline can recover such a strategy when its generating process is known.

To make the synthetic dataset resemble the empirical data as closely as possible, we used 30 board configurations that participants encountered in the behavioral experiment and generated 5 independent synthetic behavioral trajectories for each configuration, yielding 150 trajectories in total. Within each board configuration, the 5 trajectories were generated with lapse probabilities of 0.0, 0.0, 0.1, 0.1, and 0.2, respectively. These values were chosen to approximate the distribution of lapse probabilities estimated when fitting the original ColorBlocking model to participants: most participants had low estimated lapse rates, and approximately half had an estimated lapse rate of 0.0. This design therefore preserves the empirical task configurations and approximates the level of stochasticity observed in the original behavioral data while allowing the underlying strategy to be controlled.

\subsection{Generating Revised Programs}
For each synthetic behavioral trajectory, we independently ran our program refinement pipeline twice, resulting in two candidate revised programs per synthetic trial. Of the 300 generated programs, 287 were validated by the Audit Module. Among those programs, 215 had the expected program interface and could be evaluated using the same likelihood-based inference procedure as the base CB model. The remaining programs were excluded because their generated interfaces required additional arguments or otherwise could not be evaluated under the standardized procedure. Thus, the synthetic analysis reports results for 215 refined programs.

\subsection{Synthetic analysis results}
We evaluated each discovered program on 40 independently generated held-out trajectories from the same \texttt{CB\_damaged\_2} strategy and task configurations that were not included in the generation of revised programs. For each program, we compared its mean held-out log likelihood with those of the original ColorBlocking (CB) model and the generating \texttt{CB\_damaged\_2} model. Across the 215 evaluated programs, 111 (51.6\%) achieved higher mean held-out log likelihood than CB, while 2 (0.9\%) matched or exceeded the generating model. To quantify partial predictive recovery, we computed the fraction of the predictive gap between CB and CB-damaged recovered by each revised program:
$$
R = \frac{\overline{\mathrm{LL}}{\mathrm{revised}}-\overline{\mathrm{LL}}{\mathrm{CB}}}
{\overline{\mathrm{LL}}{\mathrm{CB\text{-}damaged}\text{-}2}-\overline{\mathrm{LL}}{\mathrm{CB}}}.
$$
The median recovery was 0.334. Overall, 109 programs (50.7\%) recovered a positive fraction of the predictive gap, 18 (8.4\%) recovered at least 75\%, and 2 (0.9\%) matched or exceeded the generating model. These results indicate that the discovery procedure can recover predictive consequences of a richer strategy, but recovery is heterogeneous: because revisions are generated from individual trajectories, they can also overfit trajectory-specific features and fail to generalize even when the underlying data-generating strategy is known. Importantly, predictive recovery does not imply exact recovery of the generating computational mechanism; multiple distinct implementations of damage-repair behavior were produced by the discovery procedure.

\end{document}

%% file: math_commands.tex
\usepackage{amsmath,amsfonts,bm}

\def\eqref#1{equation~\ref{#1}}

\def\1{\bm{1}}

\DeclareMathAlphabet{\mathsfit}{\encodingdefault}{\sfdefault}{m}{sl}
\SetMathAlphabet{\mathsfit}{bold}{\encodingdefault}{\sfdefault}{bx}{n}

